\pdfoutput=1

\documentclass[11pt]{article}
\usepackage[final]{emnlp2022}
\usepackage{graphicx}
\usepackage{booktabs}
\usepackage[multiple]{footmisc}
\usepackage{cleveref}
\usepackage{times}
\usepackage{latexsym}
\usepackage{hyperref}
\usepackage{multirow}

\usepackage{todonotes}

\usepackage[T1]{fontenc}
\usepackage[utf8]{inputenc}

\title{A Benchmark and Dataset for Post-OCR text correction in Sanskrit}

\author{First Author \\
  Affiliation / Address line 1 \\
  Affiliation / Address line 2 \\
  Affiliation / Address line 3 \\
  \texttt{email@domain} \\\And
  Second Author \\
  Affiliation / Address line 1 \\
  Affiliation / Address line 2 \\
  Affiliation / Address line 3 \\
  \texttt{email@domain} \\}
\author{Ayush Maheshwari$^{1}$, Nikhil Singh\thanks{~~Work done while interning at IIT Bombay.}~, Amrith Krishna$^{2}$ \and Ganesh Ramakrishnan$^{1}$ \\
\texttt{\{ayusham,ganesh\}@cse.iitb.ac.in, }\\ \texttt{\{nikhil3198,krishnamrith12\}@gmail.com}\\
$^1$Indian Institute of Technology Bombay, 
$^{2}$Uniphore
}
\begin{document}
\maketitle
\begin{abstract}
% Recent advances in deep learning have resulted in huge improvements in the performance of Optical Character Recognition (OCR) systems to produce digitized text. However, OCR systems suffer from a high word error rate for historical documents in low-resource languages. In this paper, we present a post-OCR text correction dataset in Sanskrit consisting of around 3 million words from 250 thousand sentences. The scanned documents contain around 30 books from Sanskrit literature. Further, we perform extensive experiments with state-of-the-art approaches for text correction for OCR post-correction task and report a relative improvement of more 23\% in character and word error rate. Our dataset is publicly available here.

% ==== Alternate possible abstract ====

Sanskrit is a classical language with about 30 million extant manuscripts fit for digitisation, available in written, printed or scanned-image forms. However, it is still considered to be a low-resource language when it comes to available digital resources.  In this work, we release a post-OCR text correction dataset containing around 218,000 sentences, with 1.5 million words, from 30 different books. Texts in Sanskrit are known to be diverse in terms of their linguistic and stylistic usage since Sanskrit was the `lingua franca' for discourse in the Indian subcontinent  for about 3 millennia. Keeping this in mind, we release a multi-domain dataset, from areas as diverse as astronomy, medicine and mathematics, with some of them as old as 18 centuries. Further, we release multiple strong baselines as benchmarks for the task, based on pre-trained Seq2Seq language models. We find that our best-performing model, consisting of
byte level tokenization in conjunction with phonetic encoding (Byt5+SLP1), yields a 23\% point increase over the OCR output in terms of word and character error rates. Moreover, we perform extensive experiments in evaluating these models on their performance and analyse common causes of mispredictions both at the graphemic and lexical levels. %We also perform a human-level judgement which measures the ease with which humans can improve the outputs from our best-performing model and the baseline.  
%Since a  post-OCR corrector is typically trained to handle predictions from a specific OCR, we also release a test data of 500 image that was used to predict our OCR's performance. 
%trained to capture the distributional prperties, including the mispredictions, from the OCRis dependent on the OCR predictions%Finally, we observe that these benchmark models improve their performance when used in conjunction with an OCR, fine-tuned on a subset of our own data which we release with our dataset.
Our code and dataset is publicly available at \url{https://github.com/ayushbits/pe-ocr-sanskrit}.
% \href{https://anonymous.4open.science/r/pe-ocr-sanskrit-0C6A}{here}.
\end{abstract}

%which outperforms the current state of the art in post-OCR text correction for Sanskrit. Moreover, we also release a new baseline OCR, using a subset of the data from these books, and We then additionally extensively benchmark and analyse current state of the art systems in post ocr text correction and report a relative improvement of more 23\% in character and word error rate. Moreover, we also release an improved version of the Tesseract OCR which outperforms the current OCR by XX WER. Finally, we perform a human evaluation which shows human preferences for corrections from the post ocr correction system as comapred to that of coming from only an OCR

\section{Introduction}

\begin{figure}[t]
    \centering
    \includegraphics[width=0.85\linewidth]{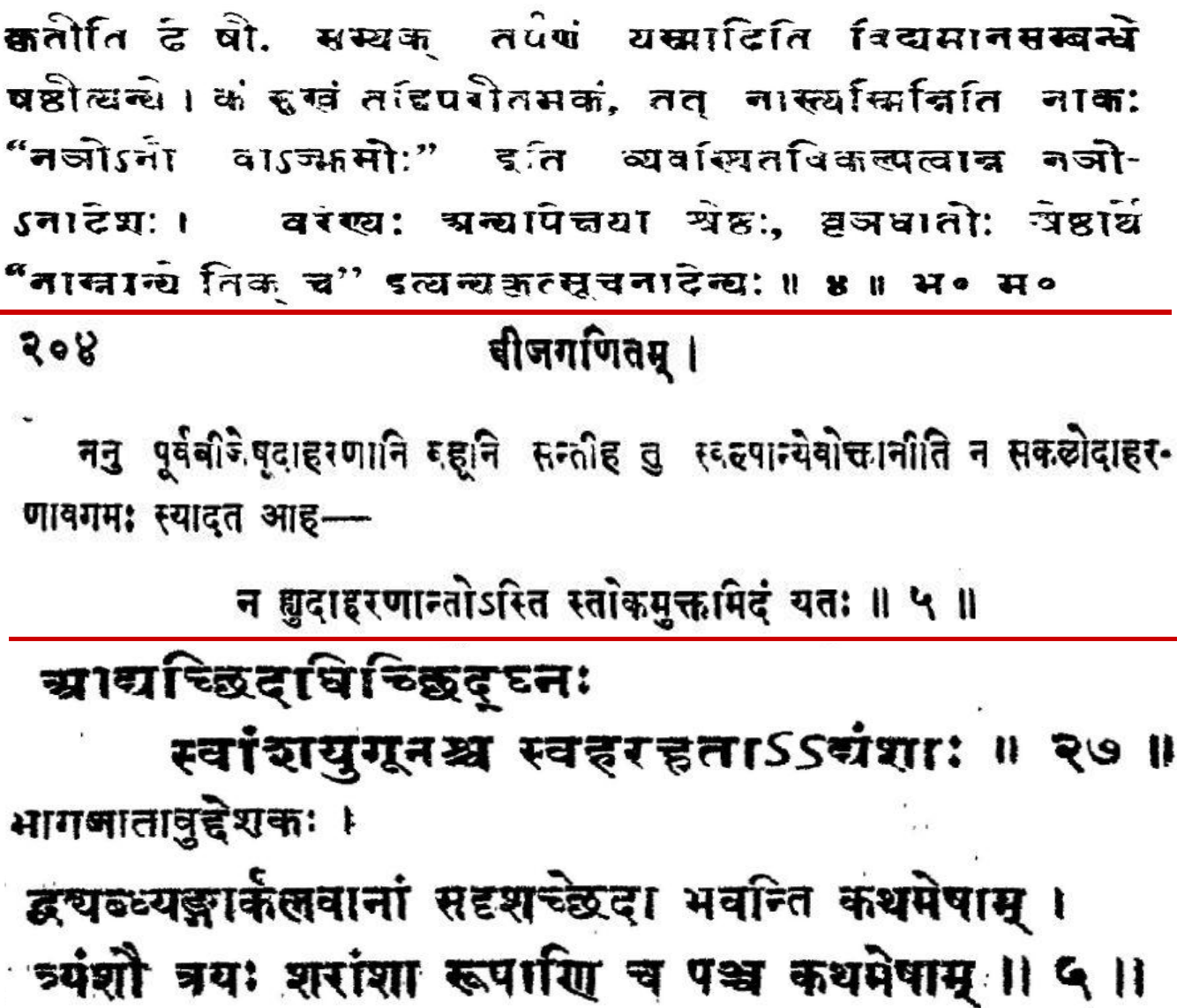}
    \caption{Image samples from different pages of our dataset.}
    \label{fig:snippets}
\end{figure}

Post-OCR text correction is a crucial post-processing step employed for correcting errors from the predictions of Optical Character Recognition (OCR) systems~\cite{rijhwani2021lexically}. A post-OCR corrector leverages the distributional information encoded in language models that aims to not only handle the systemic errors introduced by the OCR engine but also to predict meaningful and fluent sequences based on the context \cite{saluja2019ocr}. For a language like Sanskrit, the sources of OCR errors are diverse, owing to the availability of printed historical documents that vary vastly on a number of factors such as scan quality, book layout, typefaces, the orthographic similarity of letters in the alphabet, \emph{etc.} (see Figure \ref{fig:snippets}). Moreover, processing texts in Sanskrit is often challenging as the language is morphologically rich, lexically productive, follows relatively free-word order and is a low-resource language with limited available machine-readable corpora~\cite{krishnaGraph}.

In this work, we release a large Sanskrit post-correction dataset of more than 218,000 manually verified sentences, consisting of 1.5 million words. Our dataset consists of sentences from 30 books from domains as diverse as philosophy, literature, astronomy, medicine,  mathematics \emph{etc.}. Figure \ref{fig:snippets} shows a sample of the scanned images from these books, from which we obtained our dataset. Further, the sample clearly demonstrates the diversity in some of the aforementioned factors affecting the quality of the OCR predictions.   Historically, depending on the region or the time period in which it was used, several writing systems and scripts were employed for writing Sanskrit. However, the advent of the printing press largely standardized the use of the `Devan\={a}gari' script as the default writing system for Sanskrit.
% , which is presented in   Figure~\ref{fig:snippets} as well. 

We additionally release a set of strong  seq2seq baselines to benchmark for the task, including a CopyNet based LSTM model \cite{gu2016incorporating}  and four pretrained seq2seq systems (LMs). We find that all the pretrained-LM based baselines improve over the predictions from the original OCR. The best model which invokes byte level tokenization, {\em viz.}, ByT5~\cite{xue2022byt5}, in conjunction with phonetic encoding (SLP1),
% \todo{In the abstract and introduction we say
% "our best-forming model"... and one has to wait until line 245 to read for the first time what the best-performing means!
% Line 245: "The ByT5 configuration with SLP1 encoding of text currently yields the best outcome in our experiments."
% Dont you think this is a big gap in the presentation? In the abstract and intro at the least we should mention the very characterizing aspect of ByT5 and SLP1 combination that we propose and hint with rationalization that this configuration is what we propose?}
among these benchmarks (all described in \S~\ref{sec:systems}) reports a character and word error rates of 2.98\% and 23.19\% respectively, as against that of 3.89\% and 30.23\% from the original OCR. This is primarily due to the ability of byte-level tokenizers to learn arbitrarily longer text in a setting where the frequency of words is low and out-of-vocabulary words are high (\S \ref{sec:dataset}). Moreover, this goes well with the fact that the writings in  Sanskrit follow a phonemic orthography, i.e. phonemes have a direct one-to-one correspondence with the orthographic symbols. We identify that most errors arising from the original OCR are from mispredictions in word boundary detection, diacritics and orthographically similar characters. Further, as the performance of the post-OCR text correction system is highly dependent on the predictions of the OCR, we also release the test dataset used for testing our current OCR. The  test dataset consists of 500 images and their corresponding text, which can be used to benchmark an OCR, prior to using its predictions for post-OCR text correction. 

%As part of our benchmark, we also release 20,000 synthetically generated  images with their corresponding text which was originally used to train the current tesseract based OCR model.along with results from an OCR, fine-tuned on a small subset of the current dataset. 

%In addition, we also release an OCR test dataset in Sanskrit containing around 500 images and their corresponding text from different domains. Finally, we perform an extensive study on post-correction task using seq2seq models.

%owing to systemic yet diverse errors from  Optical  introduced by miltiple factors involved in Post-OCR text correction can be often challenging, especially  for a classical language like Sanskrit, where the printed historical texts can be vastly diferent in termsHowever, the task can be often challenging for a low-resource language like Sanskrit Post-OCR text correction is often challenging  

%With the advent of deep learning, have vastly improved the performance and speed of text digitization. However, performance of OCR systems is not consistent across languages and is influenced by the  of the document. The problem is severe for historical documents in low resource languages such as Sanskrit. In Figure~\ref{fig:snippets}, we show  in the \textit{devanagari} script. Hence, it is desirable to perform post-processing on the OCR output to produce an improved text.

%

\section{Dataset}\label{sec:dataset}

Sanskrit used to be the `lingua franca' for scholarly discourse in the Indian subcontinent for about three millennia and the classical language is still in sustenance in the region. It is estimated that as many as 30 million extant documents, more than that in Greek and Latin combined, are fit for digitisation in Sanskrit \cite{goyal-etal-2012-distributed, adiga2021automatic}. The current corpus is released as part of our attempt at large scale digitisation of old manuscripts in Sanskrit. Our corpus contains about 30 books, a subset of 103 books in our digitisation pipeline. These books were originally published at least a century ago, and are manually verified to have no copyright issues. We consider printed versions of these books, most of them reprinted in the first half of the twentieth century. While, these books are widely accessible to the public via libraries and academic institutions, we manually had to scan several of them as part of its digitisation process. These books vary widely in their vocabulary and stylistic usage owing to the differences in the domain and the original time period of publication, where the latter can be as old as the fifth century AD. 

We release a multi-domain dataset from 30 different books and have 218,000 manually verified sentences in it. The share of each book in the corpus amounts to 3.33\% on average with a variance of 4.09, in terms of the number of pages. %The largest book contributes to 7.71\% of the pages and the smalles one contributes amin:0.39%), in terms of the number of pages.
The corpus consists of more than 1.5 million tokens, with an average frequency of 2.59, and has a  vocabulary of 581,445 unique words. Further, 88\% has a frequency of one and more than 96\% of the words appear less than 5 times. Such a frequency distribution of tokens in Sanskrit corpora is common, given the morphological richness and lexical productivity (due to compounding) in Sanskrit~\cite{krishna-etal-2017-dataset,dcsOliver}. Further, the average word length is 10.4 characters. In Table~\ref{tab:dataset}, we present count of sentences and tokens in our train-test-val split.  Of the 20,738 words in the test data vocabulary, 54.53\% of those are out of vocabulary. Earlier approaches to post-OCR text correction have employed lexicon-driven approaches for several languages, though such approaches without a wide coverage lexicon might be challenging for a language like Sanskrit~\cite{bassil2012ocr,carlson2007memory}. 

\begin{figure}
    \centering
    \includegraphics[width=1\linewidth]{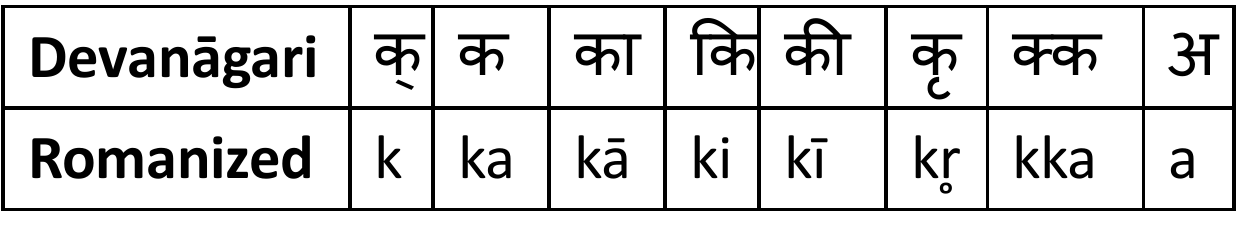}
    \caption{Devan\={a}gari and Romanised  representation of `k' followed by different vowels. Conjunct consonant (`kka') may also have separate symbols in Devan\={a}gari.}
    \label{fig:script}
\end{figure}

\begin{table}[!t]
\centering
\resizebox{0.8\linewidth}{!}{%
\begin{tabular}{|l|c|c|}
\hline
\multicolumn{1}{|c|}{\textbf{Split}} & \textbf{\# sentences} & \textbf{\# words} \\ \hline
Train                                & 208,173                & 1,444,913           \\ \hline
Validation                           & 5000                  & 34,762             \\ \hline
Test                                 & 5000                  & 34,705             \\ \hline
\end{tabular}%
}
\caption{Number of words and sentences in the dataset split.}
\label{tab:dataset}
\end{table}

Prior work in post-OCR text correction in Sanskrit \cite{krishna2018upcycle} focused on texts written in IAST or Romanized Sanskrit \cite{monier1899sanskrit}, while the current dataset is focused on Devan\={a}gari. All the books we consider here use Devan\={a}gari script as the writing system. While Devan\={a}gari is in existence since the fourth century CE \cite{bombay1896gazetteer}, it has become the primary standard for writing Sanskrit, and several other languages, with the advent of the printing press in India. The script consists of 47 primary characters and is a left-to-right abiguda, where contiguous consonant-vowel sequences are treated as unitary units.  As shown in Figure \ref{fig:script}, the vowels following the consonants (k in the figure), are treated as secondary units. These units are expressed as `m\={a}tras' in the writing system, which essentially are  diacritic markers. These markers may appear before, after, above or below an orthographic  consonant symbol as shown in Figure \ref{fig:script}. The same vowel, say `\textit{a}' in the romanised script, is written as a different character when used independently as a primary unit. Similarly, conjunct consonants, like `\textit{kka}' in the figure, would also result in a different orthographic unit, leading to an increase in possible output units for the original OCR. Moreover, these orthographically similar units can also be confusing to an OCR system. 

\subsection{OCR Editing Process}
The current work is part of an OCR project that aims to digitize hundreds of Sanskrit books present in scanned image format\footnote{Project website: \url{https://www.cse.iitb.ac.in/~ocr/}}. The dataset is an outcome of a publicly funded project, primarily carried out by researchers at IIT Bombay. The project currently has 103 books in its pipeline. Our dataset consists of books primarily from philosophy, literature, mathematics. medicine and astronomy. List of books is provided in Figure \ref{tab:books} in appendix.

%Our dataset consists of books that were originally published at least a century ago, and are manually verified to have no copyright issues. We mostly consider the printed versions of these books, most of them reprinted in the first half of the twentieth century. Moreover, these books are widely accessible to the public via libraries and academic institutions in printed and scanned versions. However, these are not digitised beyond mere scanning of the printed copies.

To aid the process of correction of OCR output, we developed an open-source post-OCR editing tool \cite{maheshwari2022udaan} that reduces the cognitive and editing load of the users and increases the speed of text correction. The in-house developed tool is used for OCR correction, verification and proofreading.  Currently, 14 experts contribute to various stages of the digitisation process. These experts are either linguists, trained specifically in Sanskrit linguistics, or computational linguists, and seven of them are working full time for the project. Each page in the book passes through a three step process, and a separate expert oversees each step. The 3 steps are: 1) Manual correction/post-editing of OCR prediction by looking at the original scanned image, 2) Verification of the corrected text performed in the previous step, and 3) Proofreading of the text to check for obvious errors. Verification is primarily aimed at maintaining fidelity of the corrected text to the scanned lines and proofreading is aimed at ensuring linguistic and semantic correctness of the text.

\section{System Descriptions} \label{sec:systems}
OCR post-correction is a text correction task which can be formalised as a monotone seq2seq model \cite{schnober-etal-2016-still}. We use an encoder-decoder framework that takes predictions from an OCR  as its input. %Encoder finds the representation of the input character sequence which are then fed to the decoder for prediction. 
While we use multiple pre-trained seq2seq models as our baselines, none of these has Sanskrit as one of their languages. However, Devan\={a}gari script is employed in other languages, such as Hindi, which are present in these models. Secondly, unicode encoding of Devan\={a}gari often poses several challenges owing to the variable byte length employed per character for encoding. Hence, we losslessly transliterate the text into SLP1, an ASCII-based case-sensitive transliteration scheme in our experiments. 
%While the pretrained language models by default employ their own subword or byte-level tokenisers, we use a su
%Sanskrit is written in \textit{devanagari} script, however, several existing multi-lingual models are not trained for \textit{devanagari} characters.  Secondly, the problem of  is present in Indic scripts. Therefore in addition to \textit{devanagari} characters, we also trained with ASCII transliteration scheme, SLP1 (Sanskrit Library Phonetic Basic) encoding format in our experiments.
%Sanskrit is a morphologically rich language where a verb may have more than 90 inflections while a noun can have 72 different inflections. Therefore, we use a Byte Pair Encoding (BPE) or character/byte-level as tokenizer in our experiments.

\noindent \textbf{Baseline OCR Model} : Our baseline OCR model is an OCR engine that uses the Tesseract OCR \cite{smith2007overview}. The model is fine-tuned upon 20,000 synthetically-created  images with the Sanskrit language flag. 
% We train Tesseract with . 
% The comparison of fine-tuned OCR with original Tesseract OCR on the test set of 500 pages is given in Table \ref{tab:ocr}. 
We release our OCR test set along with the post-OCR correction dataset.

\noindent \textbf{CopyNet} \cite{gu2016incorporating}: uses a copying mechanism %, or rather learns a copy distribution in addition to the prediction distribution, 
in an LSTM-based seq2seq framework to leverage the (partial) overlap between input and output strings. The model consists of 3 LSTM modules stacked on top of each other for both the encoder and decoder. Following~\citet{krishna2018upcycle}, we use BPE for learning the vocabulary, which has shown the ability to handle corpora with `rare words' \cite{sennrich-etal-2016-neural}.

\noindent \textbf{mBART} \cite{mbart} is a multilingual variant of the BART, both of which are seq2seq models. It has an autoregressive decoder and  a BERT-based encoder. %The noise is injected either using random span masking or reordering of the source sentence. 
%mBART is primarily used for multi-lingual translation task though it has been widely for several other seq2seq tasks including, error correction task \cite{dutta2022error}. 
We used mBART-50 \cite{tang2020multilingual}, specifically its \textit{HuggingFace} implementation (large), in our experiments which has been trained on large monolingual corpus of 50 languages. Here, we use text in its original form as well as in the transliterated SLP1 form.  %In our experiments, we use mBART large model from \textit{huggingface} with BPE tokenization and use both \textit{devanagari} and SLP1 encoded text.

\noindent\textbf{mT5} ~\cite{xue2021mt5} is a multilingual variant of T5 \cite{raffel2020exploring}, trained on 107 languages. T5 is a seq2seq text generation model, pretrained on  a mixture of supervised and unsupervised tasks using a span-corruption objective. In experiments, we employ the mT5 base model from \textit{HuggingFace} along with BPE tokenization and use both dev\={a}nagari and SLP1 encoded text.
%where every task is treated as a text generation problem. It is pre-trained on  We use a multi-lingual variant of T5, {\em viz.}, , s which uses a basic encoder-decoder architecture of original Transformer model \cite{vaswani2017attention}. 

\noindent\textbf{ByT5} \cite{xue2022byt5}: Given that we have a corpus with mostly `rare words' ({\em c.f.}, \cref{sec:dataset}), any unseen set in Sanskrit will suffer from out-of-vocabulary words. A natural solution is to tokenize words at a character level where each character is represented by UTF-8 bytes. ByT5 is a variant of mT5 except that model is fed with a fixed 256 byte values. In experiments, we use ByT5 small model from \textit{HuggingFace} with byte tokenizer and SLP1 encoded text.

\noindent\textbf{IndicBART} \cite{dabre2021indicbart}: It is a multi-lingual BART-based model trained on 11 Indic-family languages in dev\={a}nagari script. The model is roughly half the size of mT5 model.

\section{Experiments and Results}

In Table \ref{tab:postOCR}, we present the macro-averaged Word Error Rate (WER) and Character Error Rate (CER) for each of our baseline systems. The predictions directly from OCR report a CER and WER of 3.89\% and 30.23\% respectively. In our experiments, CopyNet's predictions worsen as per both  our metrics, resulting in a CER and WER of 13.25\% and 50.38\% respectively. However, all of the pre-trained language model configurations employed for post-OCR correction improves over the original OCR predictions. In general, we find that use of Devan\={a}gari scripts instead of SLP1 to encode text in Sanskrit, results in improved performance for the task. However, with ByT5, we find that our model produces truncated outputs mostly due to an increase in sequence length in ByT5 due to the byte-level vocabulary used in it. The output from ByT5-Dev has a  CER of 6.17\% and a WER of 27.72\%, higher than that of ByT5-SLP1, when a sequence length of 1024 was used. Even though the CER is further reduced to 4.59 (from 6.17) for ByT5-Dev, when its maximum sequence length is reconfigured  to 2048, it still does not outperform ByT5-SLP1 (refer Table \ref{tab:impact}). The use of SLP1 encoding for Sanskrit converts them to ASCII sequences, thereby reducing the overall sequence length for input. The ByT5 configuration with SLP1 encoding of text currently yields the best outcome in our experiments. We discuss the impact of different sequence length and memory overheads between Dev and SLP1 variants of ByT5 in Appendix \ref{sec:seqlength}. %ByT5 outperforms other models by a significant margin. % primarily due to its tokenization scheme which do not suffer from out-of-vocabulary words problem.

\begin{table}[!t]
\centering
\resizebox{\linewidth}{!}{%
\begin{tabular}{@{}cccc@{}}
\toprule
Encoding & Model     & CER           & WER            \\ \midrule
Dev      & OCR       & 3.89          & 30.23          \\ \midrule
Dev      & mBART     & 3.50 {\scriptsize{(+10)}}         & 26.11 {\scriptsize{(+13.7)}}          \\
SLP1     & mBART     & 3.71  {\scriptsize{(+4.5)}}        & 26.60 {\scriptsize{(+12)}}          \\
Dev      & IndicBART & 3.55 {\scriptsize{(+8.7)}}         & 25.73 {\scriptsize{(+14.9)}}          \\
Dev      & CopyNet   & 13.25   {\scriptsize{(-240)}}      & 50.38 {\scriptsize{(-66)}}          \\
SLP1     & mT5       & 3.53 {\scriptsize{(+9.2)}}         & 26.47 {\scriptsize{(+12.5)}}          \\
Dev      & mT5       & 3.34 {\scriptsize{(+14.1)}}          & 25.57 {\scriptsize{(+15.4)}}          \\
SLP1     & ByT5      & \textbf{2.98} {\scriptsize{(+23.4)}} & \textbf{23.19} {\scriptsize{(+23.3)}} \\ \bottomrule
\end{tabular}%
}
\caption{CER and WER (lower is better) on Post-OCR correction task for different encoding schemes and model. Numbers in brackets () corresponds to percentage improvement over OCR model output (top row). All methods are evaluated on dev\={a}nagari text and all models except ByT5 uses BPE tokenizer. Dev refers to dev\={a}nagari.}
\label{tab:postOCR}
\end{table}
We observe three primary sources of errors from the original OCR predictions, namely, word and sentence boundary identification owing to missing or extraneous prediction of space and sentence boundary markers, mispredictions due to m\={a}tra or diacritics, and mispredictions arising out of orthographically similar characters. All of these cumulatively contribute to 61.76\% of the character-level errors. Boundary detection at the word level and at the sentence level, identified by a space marker or by sentence terminating punctuation, contributes to 26.96\% of the OCR errors. 89.3\% of the boundary detection errors  arise out of identifying word boundaries. Similarly, errors due to incorrect or missing m\={a}tras (diacritics) contribute to 22.41\% of all the errors in OCR. In Sanskrit, these m\={a}tras are generally secondary vowels following a consonant, a phenomenon common in abiguda writing systems (\S \ref{sec:dataset}). Mispredictions specifically due to orthographically similar characters contribute to 12.39\% of the total errors. With ByT5, the best performing model we report, we find an error reduction of 44.45\%, 37.74\% and 14.16\% reduction in boundary identification,  m\={a}tra prediction and error corrections from orthographically similar predictions respectively. 

As a further analysis, we collect the most frequent 300 tokens in the corpus, with at least three letters for a word, and find a total of 2875 occurrences in the ground truth corpus. Among these frequent tokens, OCR predicts each of them correctly at least once. Further, to identify possible similar tokens predicted instead of the correct token, we use the Ratcliff pattern recognition algorithm \cite{black2004ratcliff}, with a matching ratio of 0.6. Here, we find that 8.27\% of the token occurrences among the most frequent tokens do not have a corresponding prediction that satisfies the matching criteria. With ByT5, this number is reduced by 2.09\%. Moreover, in both OCR predictions and ByT5 based predictions, we mostly find unique patterns in mispredictions for each token and are not able to find any consistent or systemic patterns for each token.

% \noindent\textbf{Effect of increasing sequence length on ByT5-Dev}: 

%A major source of error in the OCR, is primarily in spacing between words or finding the right word boundaries.  75.92\% of the errors incurred by the OCR, during its prediction, are in identifying the word boundaries. More specifically,  65.13\% of all the errors arise out of the OCR not identifying the word boundaries, thereby requiring a need to insert a space at the appropriate word boundary. 10.79\% of the errors in the OCR arise out of adding unnecessary space between the characters of a single word. With ByT5, there is a 21.92\% reduction in word boundary-related errors.  ByT5 further results in a 37.8\% reduction in character-related errors. 
\subsection{Experiments on out of domain test dataset} 
\begin{figure}
    \centering
    \includegraphics[width=\linewidth]{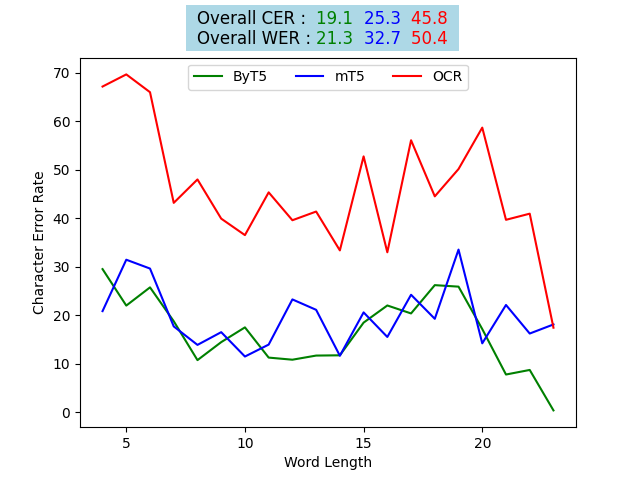}
    \caption{Comparison of CER with different word lengths on an out-of-corpus test set of 500 sentences. }
    \label{fig:word-cer}
\end{figure}

Similar to prior works \cite{rijhwani-etal-2021-lexically,krishna-etal-2017-dataset}, we ensure that there is no sentence-level (sequence) overlap between the train, test and validation split. Though there is an overlap in terms of the books, we ensure that none of the test-data sentences are seen during training. To test the generalizibility of our models, we use an out-of-domain test data comes from a completely new book, Brihat-samhita, not included in any of the train-test-validation splits. We also release a new out-of-domain test dataset which comes from a text that is not part of any of the 30 texts included in our dataset. Figure \ref{fig:word-cer} shows the performance of all the OCR, ByT5 and the mT5 systems for this dataset. Here, ByT5 has shown to significantly reduce the CER and WER from the OCR outputs. 

\section{Conclusion}

We release a dataset consisting of 218,000 sentences from 30 books for Sanskrit Post-OCR text correction. We also release a set of strong baselines as a benchmark, which currently shows consistent and significant improvements from the OCR predictions, both on the in-domain test data and out-of-domain test data. All our baselines, in spite of not seeing Sanskrit during pretraining, have shown to generalise well for the task. While using Devan\={a}gari Unicode encoding for our experiments has shown to perform better than using SLP1 for multiple baselines, SLP1-based encoding on ByT5 gives the best performance overall.

\section{Limitations}

A major limitation with the current baselines is the mispredictions happening at the word level. Here, of the  mispredicted words by ByT5-SLP, our best performing model, 71.17\% are not even valid words in Sanskrit. None of our pretrained models currently are lexically or morphologically aware resulting in the formation of invalid words in the language. Moreover, owing to the low-resource nature of the language, none of the pretrained language models we employed used Sanskrit for its pretraining. An immediate challenge with outputs of our post-OCR text correction would be the use of these predictions for downstream tasks, which are heavily reliant on rule-based morphological analysis of these words \cite{krishnaGraph}. We plan to incorporate morphologically aware self-training approaches and dynamic markup decoding \cite{decao2020autoregressive} which can incorporate various valid inflected forms of a stem in a trie to handle such scenarios.

\section{Acknowledgements}
We thank anonymous reviewers for providing constructive feedback. Ayush Maheshwari is supported by a Fellowship from Ekal Foundation (www.ekal.org). Ganesh Ramakrishnan is grateful to NLTM OCR Bhashini project as well as the IIT Bombay Institute Chair Professorship for
their support and sponsorship.

%Another issue we currently face is the increased training time due to the increased sequence length in ByT5, our best-performing model. 
%encoding text in Sanskrit     %To the best of our knowledge, this is the second biggest collection of distinct Sanskrit texts (in terms of sentences), second only to DCS \cite{dcsOliver}. 

\bibliography{refs}
\bibliographystyle{acl_natbib}
\newpage
\newpage
\newpage
\appendix

\begin{center}
{\huge{Appendix }}    
\end{center}

\section{Impact of difference sequence length}\label{sec:seqlength}

ByT5 splits each character into bytes. Since the unicode encoding of Devan\={a}gari characters typically have higher byte lengths, an input sequence in ByT5 often tends to be shorter, affecting contextual information in longer sentences. We present corresponding results in Table \ref{tab:impact}.

% \usepackage{graphicx}
\begin{table}[]
\centering
\resizebox{\linewidth}{!}{%
\begin{tabular}{|c|cc|cc|}
\hline
Max token & \multicolumn{2}{c|}{2048} & \multicolumn{2}{c|}{1024} \\ \hline
\textbf{Model} & \multicolumn{1}{c|}{\textbf{CER}} & \textbf{WER} & \multicolumn{1}{c|}{\textbf{CER}} & \textbf{WER} \\ \hline
OCR & \multicolumn{1}{c|}{4.02} & 30.7 & \multicolumn{1}{c|}{4.67} & 30.02 \\ \hline
ByT5-Dev & \multicolumn{1}{c|}{4.59} & 26.6 & \multicolumn{1}{c|}{13.12} & 38.6 \\ \hline
ByT5-SLP1 & \multicolumn{1}{c|}{\textbf{3.05}} & \textbf{23.5} & \multicolumn{1}{c|}{\textbf{3.49}} & \textbf{25.3} \\ \hline
\end{tabular}%
}
\caption{ByT5-Dev and SLP1 models are trained with different maximum token length.  Max token size of 2048 is equivalent to 135 characters while max token size of 1024 is equivalent to 56 characters. We truncated the maximum character length in the test set corresponding to each experiment.}
\label{tab:impact}
\end{table}

\section{List of Books}

We present list of books used in our dataset in Table \ref{tab:books}.
\begin{table*}[t]
% \centering
\resizebox{.95\textwidth}{!}{%
\begin{tabular}{|l|l|}
\hline
\multicolumn{1}{|c|}{\textbf{Book Name}} & \textbf{Genre} \\ \hline
Uttararamacharita by Bhavabhuti - commentary by Veeraraghava & Arts \\ \hline
Grahalaghava of Ganesh Daivajna & Astronomy \\ \hline
Suryasiddhanta of Ranganatha & Astronomy \\ \hline
Mahabhaskariyam & Astronomy \\ \hline
Aryabhatiya Bhashya of Gargyakerala Nilakantha Samabasiva Sastri K. Vol 1 & Astronomy \\ \hline
Aryabhatiya Bhashya of Gargyakerala Nilakantha Samabasiva Sastri K. Vol 2 & Astronomy \\ \hline
Aryabhatiya Bhashya of Gargyakerala Nilakantha Samabasiva Sastri K. Vol 3 & Astronomy \\ \hline
Karana-Kutuhalam & Astronomy \\ \hline
LaghuManasa & Astronomy \\ \hline
Aryabhatiya commentary by Suryadeva Yajvan & Astronomy \\ \hline
Khandakhadyaka & Astronomy \\ \hline
Ganak Tarangini & Astronomy \\ \hline
Bijaganita with Navankjura-Apte & Mathematics \\ \hline
Bijaganitavatamksa of Narayan Shukla & Mathematics \\ \hline
Bijaganita with Bijankura & Mathematics \\ \hline
Ganitakaumudi of Narayanapandita (vol. 1) & Mathematics \\ \hline
Rekhaganita of Jagannatha Vol. 2 & Mathematics \\ \hline
Bijaganita by Tr Abhyankar & Mathematics \\ \hline
Laghubhaskariyam Part 2 & Mathematics \\ \hline
Rekhaganita of Jagannatha Vol. 1 & Mathematics \\ \hline
Lilavati with kriyakramakri & Mathematics \\ \hline
Patiganita of Sridhara & Mathematics \\ \hline
Brahmasphutasiddhanta of Brahmagupta & Mathematics \\ \hline
Laghubhaskariyam & Mathematics \\ \hline
Hathayogapradipika by Svatmarama & Medicine \\ \hline
Mimamsanyayaprakasha by Aapdeva & Philosophy \\ \hline
Shastradipika by Parthasarthi & Philosophy \\ \hline
Shabdashaktiprakashika by Jagdishtarkalankara & Philosophy \\ \hline
Prakaranapanchika-Shalikanatha & Philosopy \\ \hline
\end{tabular}%
}
\caption{List of books used in the experiments.}
\label{tab:books}
\end{table*}

\end{document}